\documentclass[runningheads]{llncs}
\usepackage{graphicx}
\begin{document}
\title{Fully Automated Pancreas Segmentation with Two-stage 3D Convolutional Neural Networks}
%
\author{Ningning Zhao \and Nuo Tong \and Dan Ruan \and Ke Sheng}
%
\authorrunning{Authors}
%
\institute{University of California Los Angeles, School of Medicine, California, USA}

\maketitle             
\begin{abstract}
Due to the fact that pancreas is an abdominal organ with very large variations in shape and size, automatic and accurate pancreas segmentation can be challenging for medical image analysis. In this work, we proposed a fully automated two stage framework for pancreas segmentation based on convolutional neural networks (CNN). In the first stage, a  U-Net is trained for the down-sampled 3D volume segmentation. Then a candidate region covering the pancreas is extracted from the estimated labels. Motivated by the superior performance reported by renowned region based CNN, in the second stage, another 3D U-Net is trained on the candidate region generated in the first stage. We evaluated the performance of the proposed method on the NIH computed tomography (CT) dataset, and verified its superiority over other state-of-the-art 2D and 3D approaches for pancreas segmentation in terms of dice-sorensen coefficient (DSC) accuracy in testing. The mean DSC of the proposed method is 85.99\%. 

\end{abstract}

\section{Introduction}
Automated and accurate organ segmentation is a fundamental step in medical image analysis, computer assisted diagnosis and radiation therapy plans. Recently, deep learning based methods such as convolutional neural networks (CNN) have demonstrated to be powerful tools for organ segmentation thanks to the availability of large annotated datasets and computational resources compared with traditional segmentation techniques.

One issue in organ segmentation is whether to deal with 2D slices or 3D volumes since training models on both 2D and 3D scans have their advantages and disadvantages. Specifically, training models on 3D volumes directly can leverage the inherent spatial and anatomical information in volumetric organs with the cost of significantly higher computational power and memory than training 2D models. On the contrary, there are usually more training samples for 2D network training by slicing volumes in three orthogonal planes (sagittal, coronal and transverse), which sacrifices the 3D geometric information. Moreover, the fusion of 2D segmentation results to construct 3D mask is necessary. Existing deep learning based techniques for pancreas segmentation include both cases. For example, 2D networks were explored in \cite{Yu2018CVPR}, \cite{zhou2017fixed}. 3D network for pancreas segmentation were studied in \cite{Oktay2018AttentionUnet}, \cite{Zhu3D2018}. In \cite{Li2019arxiv} and \cite{VFNet2018}, the authors combined 2D and 3D networks for pancreas segmentation. 

In addition, segmentation of the small, soft and flexible organs like pancreas automatically and accurately can be difficult due to its large variations in shape, size and the varying surrounding contents in comparison with the large organs (e.g., liver, kidney, stomach, etc.). Thus, much more accurate segmentation can be achieved by using smaller input region around the target. The coarse-to-fine multi-stage techniques have been explored widely to address this problem. The basic idea is to determine the regions of interest (ROIs)/candidate regions in a coarse step followed by refining the segmentation on the ROIs. However, the ROI generation through the bounding box estimation for pancreas can be difficult. Several methods to generate meaningful regions on 2D slices with recall of value 99\% have been explored. For example, machine learning based techniques are implemented in \cite{Farag2017TMI}, \cite{DeepOrgan2015} for candidate region generation through bounding box regression. A fixed-point algorithm during testing stage was studied in \cite{zhou2017fixed}. Recurrent neural networks were considered in \cite{Yu2018CVPR} to keep the consistency between training and testing stages. \cite{Zhu3D2018} proposed to generate candidate regions using patch-based method. 

In this work, we proposed a two-stage method for automated pancreas segmentation on 3D computed tomography (CT) scans, which contains two steps: i) coarse segmentation on down-sampled 3D volumes for candidate region generation; ii) to refine the pancreas segmentation on smaller regions-of-interest (ROIs) at the finest resolution scale. The performance of the proposed algorithm was demonstrated on the NIH dataset.

\section{Method}
We denote the 3D CT scans as $X$ with size $W\times H\times D$. The down-sampled CT scans is $X^R$, where the superscript letter $R$ is the decimation factor. The ground truth masks corresponding to the original and down-sampled CT scans are represented by $Y$ and $Y^R$. $N = W\times H\times D$ is the total voxel number in the 3D scans. The vector version of the 3D volumes is denoted by their corresponding lower-case letters. For example, $y$ is the vector version of ground truth mask $Y$. The two steps of the proposed method are detailed as follows.

\subsection{Coarse scale segmentation}
Due to the high dimensionality of the original 3D CT scans, training models on the original CT scans leads to high cost of the computational power and memory, which limits the depth and architecture of the networks. Thus, we first train a 3D U-Net on the down-sampled volume with a decimation factor $R=4$. Based on the tight relationship between segmentation and localization, a candidate region of the pancreas can be extracted after obtaining the coarse scale segmentation mask. 
Note that the normalized bounding box of pancreas in down-sampled volume and original volume are the same since down-sampling operation cannot change the shape and location of the pancreas. Besides, the candidate region generation is conducted on 3D volumes instead of 2D slices since the location of pancreas on 3D volumes of different subjects are more consistent than that on 2D slices inter and/or intra subjects. 
\begin{figure}[h!]
\begin{center}
\includegraphics[scale=0.18]{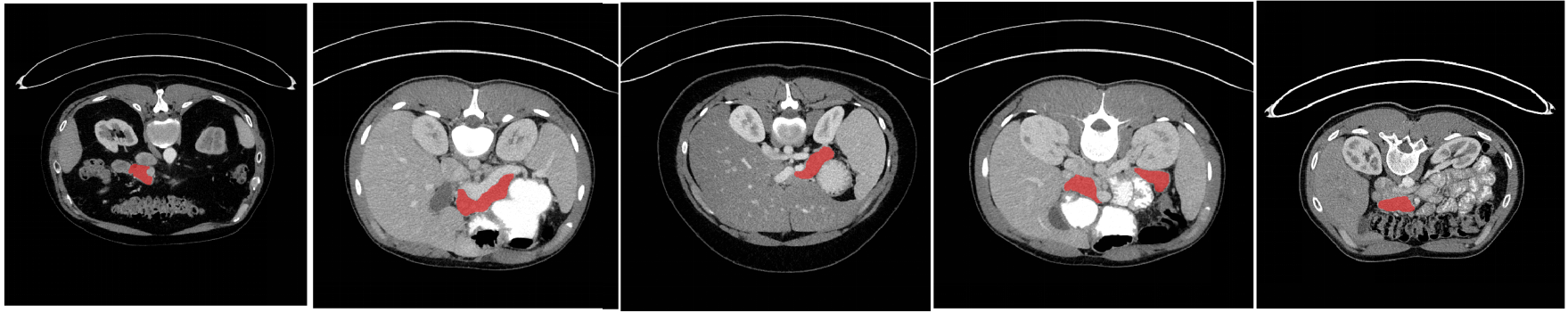}
\end{center}
\vspace{-0.5cm}
\caption{CT slices from different subjects. The anatomical properties of pancreas (indicated by red mask)  in different slices are quite different.}
\end{figure} 

\vspace{-0.6cm}
\subsection{Fine scale segmentation}
In the second stage, another UNet of the same architecture is trained on the candidate regions generated in the first stage at the finest resolution scale. Since we cannot make sure that the ROIs generated in the first stage has both high precise and recall, the second stage during training and testing procedure are implemented in different ways. 

\subsubsection{In training,} the bounding boxes are extracted from the ground-truth mask $Y$, and then enlarged by adding margins (10 pixels) along the three orthogonal axes. The candidate regions cropped with the enlarged ground truth bounding boxes (denoted as $B_y$) from the original 3D CT scans are the inputs of segmentation network in the second stage. Note that it is possible to train the networks of the two stages simultaneously since the ROIs used here are generated without the aid of the output in the first stage.

\begin{figure}[h!]
\begin{center}
\includegraphics[scale=0.18]{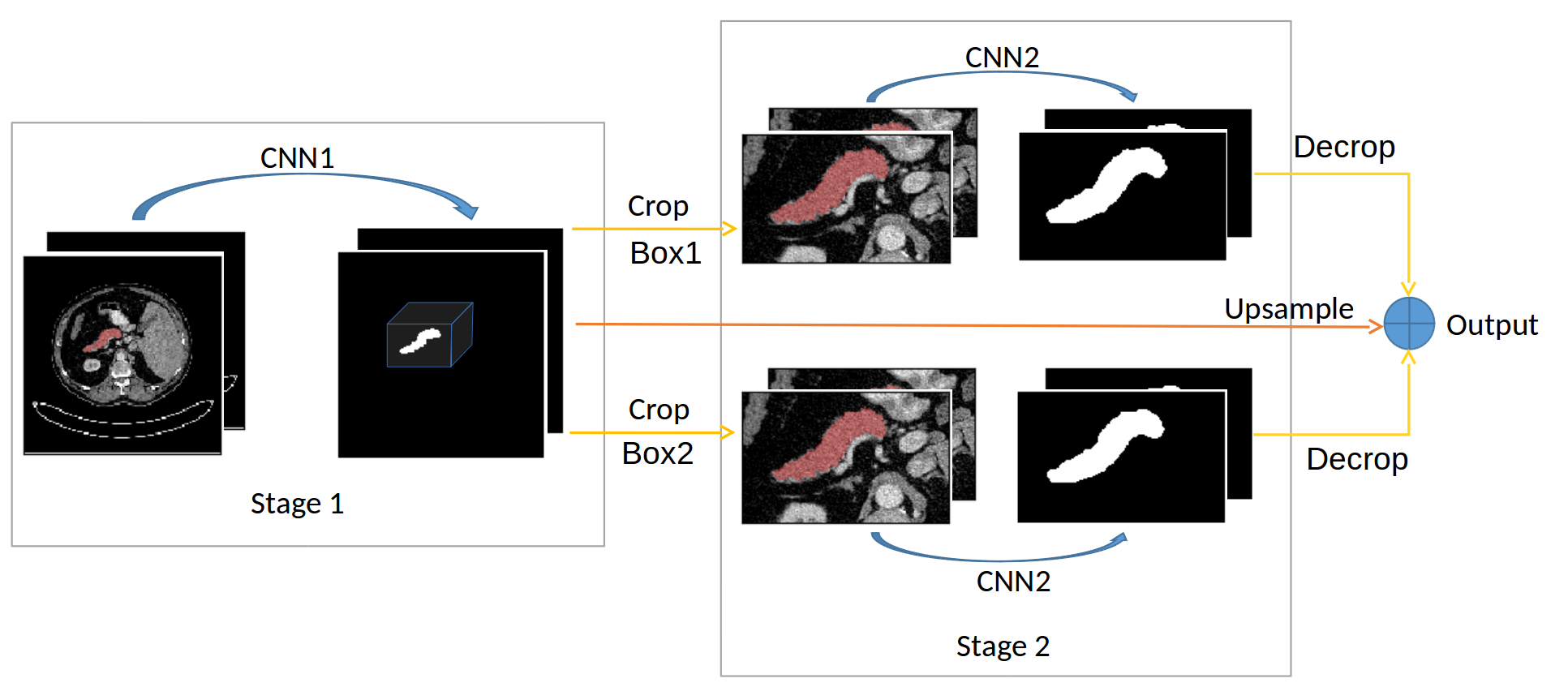}
\end{center}
\vspace{-0.5cm}
\caption{Illustration of testing procedure. In Stage 2, three channels for pancreas segmentation are used. For the two channels represented by yellow arrows, two cropped regions are fed into the same network to produce two binary masks. In the channel represented by orange arrow, the estimated mask was up-sampled with a factor R=4. The estimated mask in the three channels are finally combined through marginal voting.}
\label{testing}
\end{figure} 

\subsubsection{In testing,} after obtaining the estimated label map $\hat{Y}^R$ from the down-sampled data $X^R$, we generated two bounding boxes in different ways: i) We extracted a bounding box from $\hat{Y}^R$ directly and then enlarged it by adding margins of 2 pixels along different axes, denoted as $B_{\hat{y}}^R$. A bounding box covering the pancreas on the original CT scans was finally obtained by rescaling $B_{\hat{y}}^R$ through multiplying by R along different axes. ii) After up-sampling $\hat{Y}^R$ with a factor R, one bounding box was extracted from it directly. The bounding box was then enlarged by adding 10 pixels of margin along different axes. Note that the differences between the two bounding boxes come from the errors of sampling operation and different margins considered.  

Since two bounding boxes are generated in the testing procedure, two estimated masks can be obtained by feeding the two cropped ROIs into the U-Net separately. Finally, the two estimated masks and the up-sampled pancreas mask estimated in the first stage were combined by marginal voting. Fig. \ref{testing} explains the testing procedure of the proposed method.   

\subsection{Network architecture and loss function} 
\begin{figure}[h!]
\begin{center}
\includegraphics[scale=0.12]{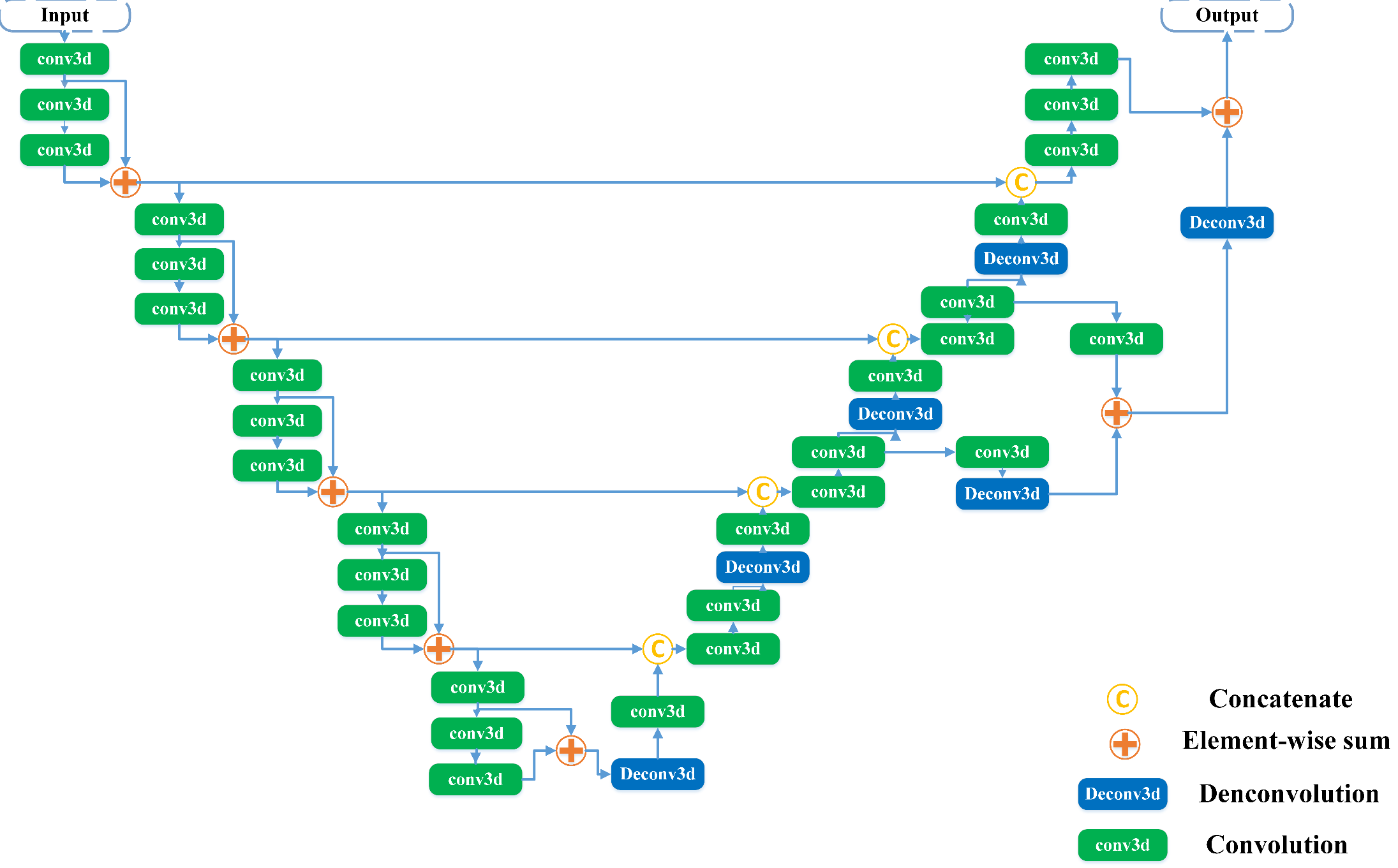}
\end{center}
\vspace{-0.5cm}
\caption{The deep network architecture. The architecture of the U-Net used in the two stages are the same.}
\label{unet}
\end{figure}

The architecture of the UNet employed in this work is shown in Fig. \ref{unet}. Note that the the architecture in stage 1 and stage 2 are the same. The parameters of the two networks in Stage 1 and Stage 2 can be trained simultaneously since they are independent during training. 

The loss function is formulated as below. Note that the same loss function are used in both stages.
\begin{equation}
\mathcal{L} = \mathcal{L}_{\textrm{Dice}} + \gamma \mathcal{L}_{\textrm{Center}}
\end{equation}
where $\gamma$ is the penalty parameter. The dice loss and center loss are given by
\begin{eqnarray}
\mathcal{L}_{\textrm{Dice}} &=& -\frac{\sum_{i}^N y_i \hat{y}_i}{\sum_i^N y_i + \hat{y}_i} - \frac{\sum_{i}^N (1-y_i) (1-\hat{y}_i)}{\sum_i^N 2 - y_i - \hat{y}_i} \\
\mathcal{L}_{\textrm{Center}} &=& \sum_{d=1}^{D} |f(Y_d) - f(\hat{Y}_d)|
\end{eqnarray}
where
\begin{equation}
f(Y_d) = (c_{x,d}, c_{y,d}) = \frac{1}{\zeta} \sum_{v=1}^H \sum_{u=1}^W (u, v) \cdot Y_d(u, v)
\end{equation}
where $f(Y_d)$ is the center point of $d$th slice of mask $Y$, which is denoted as  the weighted sum of coordinates. $\zeta = \sum_{v=1}^H \sum_{u=1}^W (u, v) Y_d(u, v)$ is the spatial normalization factor. Note that both the dice loss and center loss are differentiable. In this work, the parameter $\gamma$ is fixed as $10^{-3}$ for the initial 50 epochs and is decreased to 0 in the following epochs.

\section{Experiments}
\subsection{Dataset}
Our method is evaluated on the public NIH pancreatic segmentation dataset\footnote{https://wiki.cancerimagingarchive.net/display/Public/Pancreas-CT} \cite{DeepOrgan2015}. There are 82 contrast-enhanced abdominal CT scans and corresponding annotated labels in this dataset. The CT scans have resolutions of  $512\times512$ pixels with varying pixel sizes and slice thickness between $1.5-2.5$ mm. The pixel values for all the CT scans were clipped to $[-100, 240]$ HU, then rescaled to the range $[0, 1]$. Following previous work of pancreas segmentation, we used 4-fold cross-validation to assess the robustness of the model, i.e., 20 subjects are chosen for validation in each fold.  
\subsection{Quantitative Assessment Metrics}
In both steps, the dice similarity coefficient (DSC) is employed for segmentation accuracy evaluation. Moreover, recall and intersection-over-union (IoU) are used to assess the localization performance. The metrics used in this work are expressed as below. 
\begin{eqnarray}
\textrm{DSC} &= \frac{2 \sum_i y_i \hat{y}_i }{\sum_i y_i + \sum_i \hat{y}_i } \\
\textrm{Recall} &= \frac{\textrm{Area}(B_y \cap B_{\hat{y}})}{\textrm{Area}(B_y)} \\
\textrm{IoU} &= \frac{\textrm{Area}(B_y \cap B_{\hat{y}})}{\textrm{Area}(B_y \cup B_{\hat{y}})} 
\end{eqnarray}
where $y$ and $\hat{y}$ are the ground truth and estimated masks. $B_y$ and $B_{\hat{y}}$ are the bounding boxes extracted from the ground truth and estimated masks. 

\subsection{Results}
\begin{figure}
\begin{center}
\includegraphics[width=0.9\linewidth]{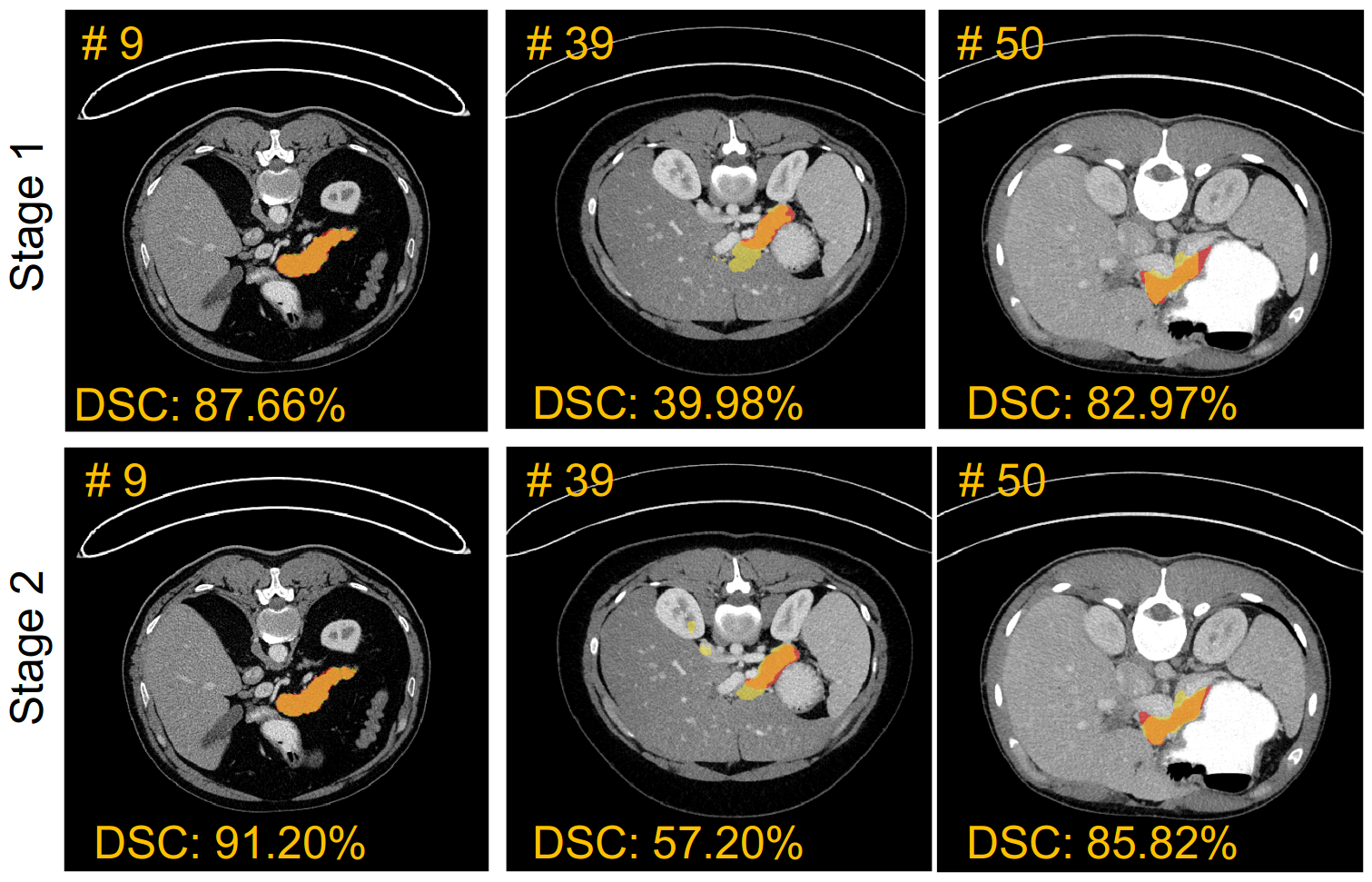} 
\caption{Examples of segmentation results of the proposed method from subjects \#9, \#39 and \#50. Red and yellow mask indicate the ground truth, prediction regions respectively. Best viewed in color.}
\label{fig_pred}
\end{center}
\end{figure}

Fig. \ref{fig_pred} shows the segmentation results from stage 1 and stage 2 respectively of three subjects in testing. For all the subjects, the overlaps between the estimated masks and ground truth masks in the second stage are larger compared to the overlaps in stage one. Quantitative results of the two-stage method are reported in Table \ref{tab1}. The mean DSC from stage 2 segmentation increases by at least 3\%, compared to the stage 1 result. 

The bounding box accuracy evaluated by recall and IoU is summarized in Table \ref{tab2}. The mean recalls of both bounding boxes are higher than 97\%. The IoUs of the two bounding boxes from different subjects are all higher than 60\% except for one subject whose IoU is lower than 40\%. However, the segmentation performance of this subject also benefits from the two-stage method, i.e., DSC increased from 39.98\% to 57.20\%. 

The comparison of pancreas segmentation using different methods are reported in Table \ref{tab_comp}. The proposed method outperforms the others in terms of the mean DSC. 
 
\begin{table}
\caption{Pancreas segmentation accuracy of the proposed method on the NIH dataset.}\label{tab1}
\vspace{-0.2cm}
\begin{center}
\begin{tabular}{|c||c|c|c||c|c|c|}
\hline 
 &\multicolumn{3}{c||}{Stage 1} & \multicolumn{3}{|c|}{Stage 2} \\
\hline \hline
Fold &  Mean DSC  & Max DSC & Min DSC &  Mean DSC  & Max DSC & Min DSC\\
\hline \hline
F0 & 82.18\% $\pm$ 5.28\% & 89.31\% & 65.59\% & 85.82\% $\pm$ 4.58\%& 91.20\% & 74.95\%\\
F1 & 78.20\% $\pm$ 10.60\% & 88.67\%& 39.98\%& 84.85\% $\pm$ 6.75\% & 90.80\% & 57.20\% \\
F2 & 83.23\% $\pm$ 3.70\% & 89.61\%& 76.36\% & 86.52\% $\pm$ 2.56\% &91.14\%& 82.00\%\\
F3 & 81.91\% $\pm$ 6.84\% &88.68\% & 78.69\% & 86.79\% $\pm$ 2.43\%& 90.64\% & 82.13\% \\
\hline
\end{tabular}
\end{center}
\end{table}

\vspace{-0.5cm}
\begin{table}
\caption{Pancreas localization accuracy of the proposed method on the NIH dataset.}\label{tab2}
\vspace{-0.5cm}
\begin{center}
\begin{tabular}{|c||c|c|c||c|c|c|}
\hline 
 &\multicolumn{3}{c||}{Bounding box 1} & \multicolumn{3}{|c|}{Bounding box 2} \\
\hline \hline
Fold &  Mean Recall  & Max Recall & Min Recall &  Mean Recall  & Max Recall & Min Recall\\
\hline
\hline
F0 & 98.44\% $\pm$ 2.23\% & 100\% & 90.24\%     & 99.41\% $\pm$ 1.69\% & 100\% & 93.22\%\\
F1 & 98.28\% $\pm$ 2.24\% & 100\% & 90.74\%     & 99.38\% $\pm$ 1.39\% & 100\% & 93.89\%\\
F2 & 97.67\% $\pm$ 4.18\% & 100\% & 85.24\%     & 98.71\% $\pm$ 3.13\% & 100\% & 88.59\%\\
F3 & 98.42\% $\pm$ 1.95\% & 100\% & 93.42\%  & 99.61\% $\pm$ 0.73\% & 100\% & 97.78\%\\
\hline
\hline
Fold &  Mean IoU  & Max IoU & Min IoU &  Mean IoU  & Max IoU & Min IoU\\
\hline
\hline
F0 & 81.04\% $\pm$ 5.18\% & 88.23\% & 69.55\%   & 69.52\% $\pm$ 5.39\% & 77.68\% & 60.04\% \\
F1 & 76.93\% $\pm$ 13.37\% & 86.78\% & 20.80\%  & 66.45\% $\pm$ 11.60\% & 77.66\% & 18.35\% \\
F2 & 77.74\% $\pm$ 4.33\% & 84.85\% & 69.43\%   & 66.78\% $\pm$ 3.85\% & 72.61\% & 59.19\% \\
F3 & 76.30\% $\pm$ 12.69\% & 90.87\% & 42.42\%  & 66.96\% $\pm$ 10.15\% & 83.38\% & 40.51\% \\
\hline
\end{tabular}
\end{center}
\end{table}

\vspace{-0.5cm}
\begin{table}
\caption{Evaluation of different methods on the NIH dataset.}\label{tab_comp}
\vspace{-0.5cm}
\begin{center}
\begin{tabular}{|c||c|c|c|}
\hline
Method &  Mean DSC  & Max DSC & Min DSC\\
\hline
\hline
Roth et.al. MICCAI'2016 \cite{HNN2016MICCAI} &  {78.01\% $\pm$ 8.20\%} & 88.65\% & 34.11\%\\
Holistically Nested 2D FCN \cite{Holger2017MIA} & 81.27\% $\pm$ 6.27\% &88.96\% &50.69\% \\
Zhou et.al. MICCAI'2017 \cite{zhou2017fixed} &  82.65\% $\pm$ 5.47\% & 90.85\% & 63.02\%\\
Attention-UNet \cite{Oktay2018AttentionUnet} & 83.10\% $\pm$ 3.80\%  & -&-\\
ResDSN C2F \cite{Zhu3D2018} & 84.59\% $\pm$ 4.86\% &\textbf{91.45\%} & \textbf{69.62\%} \\
Ours (Coarse)& 81.91\% $\pm$ 6.84\%& 89.61\%& 39.98\%\\
Ours (Refine)  & \textbf{85.99\% $\pm$ 4.51\%}& 91.20\%& 57.20\%\\
\hline
\end{tabular}
\end{center}
\end{table}
\vspace{-0.5cm}
\subsection{Discussion}
According to the quantitative results about bounding box estimation in Tab. \ref{tab2}, it is difficult to achieve high IoUs for different subjects. Thus, a new testing method different from the training procedure is introduced. In testing, two candidate regions were extracted with the estimated mask in the first stage. Then, an up-sampled segmentation mask in the fist stage and two refined segmentation masks by majority voting. Although the inconsistency between training and testing, the proposed method has achieved competitive segmentation accuracy compared with state-of-the-art algorithms. 

It is also interesting to note that more isolated false positive (FP) errors are introduced in the second stage segmentation for subject $\sharp$39. The FP error is caused by the center loss term used for training. We noticed that the center loss term contribute to increase the convergence speed during training procedure. However, it can cause more isolated false positive errors.

\section{Conclusions}
A two-stage pancreas segmentation method was proposed in this work. Two deep networks of the same architecture were trained with down-sampled and original 3D CT scans for the purpose of coarse ROI definition and refined segmentation. We also proposed a novel testing framework, which can easily used for other small organ segmentation. The proposed method has achieved competitive segmentation accuracy compared with state-of-the-art algorithms. 
\vspace{-0.3cm}
\bibliographystyle{splncs04}
\bibliography{library}

\end{document}